\documentclass{article}
\usepackage{spconf,amsmath,graphicx}
\usepackage{times}
\usepackage{epsfig}
\usepackage{graphicx}
\usepackage{amsmath}
\usepackage{amssymb}
\usepackage{bbding}
\usepackage{multirow}
\usepackage[pagebackref=true,breaklinks=true,colorlinks,bookmarks=false]{hyperref}
\usepackage{amsfonts,amssymb}

\title{Dynamic Binary Neural Network by learning channel-wise thresholds}
\name{Jiehua Zhang$^{1}$ \qquad Zhuo Su$^{1}$ \qquad Yanghe Feng$^{2}$\qquad Xin Lu$^{2}$ \qquad Matti Pietikäinen$^{1}$ \qquad Li Liu$^{2,1}$}
  
  \address{$^{1}$ CMVS, University of Oulu \\
      $^{2}$ National University of Defense Technology}
\begin{document}
%
\maketitle
\begin{abstract}
Binary neural networks (BNNs) constrain weights and activations to +1 or -1 with limited storage and computational cost, which is hardware-friendly for portable devices. Recently, BNNs have achieved remarkable progress and been adopted into various fields. However, the performance of BNNs is sensitive to activation distribution. The existing BNNs utilized the Sign function with predefined or learned static thresholds to binarize activations. This process limits representation capacity of BNNs since different samples may adapt to unequal thresholds. To address this problem, we propose a dynamic BNN (DyBNN) incorporating dynamic learnable channel-wise thresholds of Sign function and shift parameters of PReLU. The method aggregates the global information into the hyper function and effectively increases the feature expression ability. The experimental results prove that our method is an effective and straightforward way to reduce information loss and enhance performance of BNNs. The DyBNN based on two backbones of ReActNet (MobileNetV1 and ResNet18) achieve 71.2$\%$ and 67.4$\%$ top1-accuracy on ImageNet dataset, outperforming baselines by a large margin (\emph{i.e.}, 1.8$\%$ and 1.5$\%$ respectively).
\end{abstract}
\section{Introduction}
\label{sec:intro}

With the great progress made in deep learning in recent years, convolutional neural networks (CNNs) have achieved state-of-art performance in a broad range of fields. However, the existing CNNs require massive computation and storage resources to achieve high performance, which is not hardware-friendly to resource-limited devices. The Binary neural networks (BNNs), also known as 1-bit CNN, has two key advantages: 1) it constrains weights and activations to +1 or -1 to achieve a 32x memory compression ratio; 2) it utilizes XNOR and PopCount operations to replace computationally expensive multiply-add, providing a 58x computational reduction on a CPU \cite{survey2020, project2019, EBP2014, reactnet2020, birealnet2018, bnn2016}. Due to these attractive characteristics of BNNs, it has been regarded as one of the most essential neural network compression methods to have the potential for direct deployment on next-generation hardware \cite{reactnet2020}.

Despite these advantages of BNNs, the enormous accuracy gap exists between BNNs and real-valued CNN since binarized operation limits model capacity and leads to a significant information loss of feature maps. To address this problem, Bi-RealNet \cite{birealnet2018} introduced a shortcut operation to increase value range of network, achieving remarkable performance improvement compared with XNOR-Net \cite{xnornet2016}. Based on this, Liu \emph{et al.} \cite{reactnet2020} proved that BNNs are sensitive to activation distribution shift. A small distribution value offset close to zero would cause the binarized feature map to have a distinct appearance, leading to significant degradation of predictive accuracy. They proposed a generalization of Sign function (RSign) and PReLU function (RPReLU) to shift the distribution of feature maps. The proposed ReActNet based on MobileNetV1 \cite{mobilenet2017} achieved the top-1 accuracy closed to real-valued network on ImageNet dataset \cite{imagenet2015} with lower computational cost.
\begin{figure}[]
    \centering
    \includegraphics[width=\linewidth]{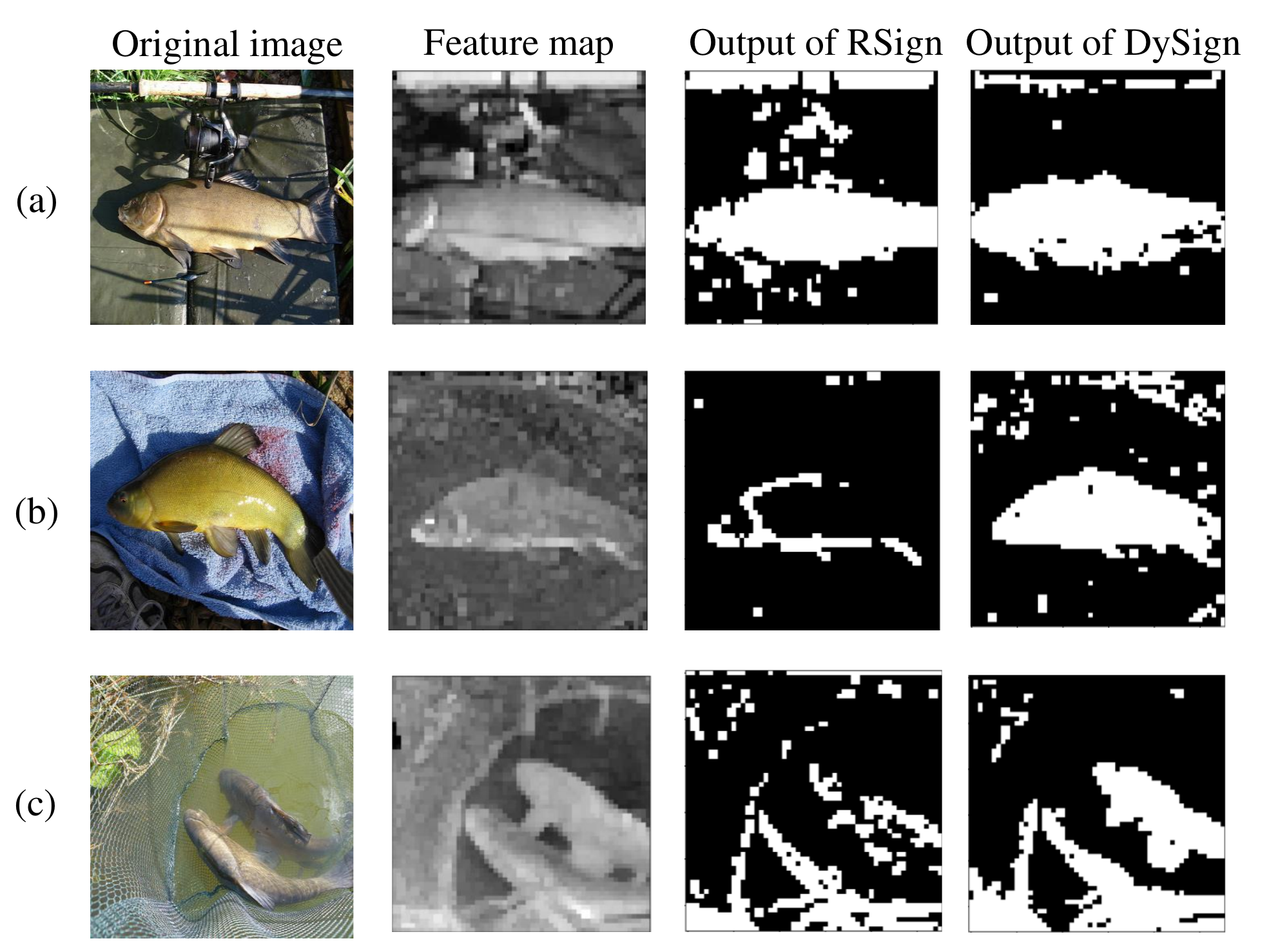}
    \caption{The output of different samples after RSign and $DySign$. A set of fixed thresholds in RSign is insufficient to diverse samples. For (a), the fixed threshold can retain the feature information of object. For (b) and (c), the feature information significantly decrease.}
    \label{fig:Fig1}
\end{figure}

However, the Sign functions in these methods apply predefined or learned \textbf{static} thresholds for adjusting activation distribution. They process the different inputs in a fixed way. The distribution of samples exists a huge gap, which means fixed shift parameters are incapable of adapting to a broad range of samples. Shown as in Figure \ref{fig:Fig1}, the fixed threshold fits with image (a) but leads to the disastrous information loss for image (b) and (c). Inspired by \cite{dyrelu2020}, we introduce dynamic learnable shift parameters based on the input feature distribution for BNNs to enhance the feature expression ability.  

Our contributions are summarized as follows:

\begin{itemize}
    \item We propose Dynamic Binary Neural Network (DyBNN) to generate diverse shift parameters based on feature distribution of the input itself. The DyBNN incorporates channel-wise \textbf{dynamic} learnable thresholds of Sign function ($DySign$) and shift parameters of PReLU ($DyPReLU$). These parameters are both learned by a dynamic learning branch consisting of a global average pooling layer and two fully-connected layers. The global information of each channel is aggregated into the hyper function and generates a set of learning shift parameters to enhance the feature expression ability.
   \item We demonstrate the effectiveness of DyBNN on the ImageNet dataset. Without bells and whistles, simply replacing static RSign and RPReLU with $DySign$ and $DyPReLU$ in two networks (ReActNet and ReActNet based on ResNet18) achieves 71.2$\%$ and 67.4$\%$ top1-accuracy, outperforming strong baselines 1.8$\%$ and 1.5$\%$ by a large margin, respectively.
\end{itemize}
\section{Methodology}
In this section, we first introduce the BNNs and then present the influence of fixed thresholds in Sign function. Finally, we illustrate our DyBNN how to reduce the information loss and enhance the model expression ability. 
\subsection{Preliminary}
For the convolution process in the BNNs, we denote the weights and features in the \textsl{l}-layer as $W^l$ and $F^l$. The input of the \textsl{l+1}-layer can be expressed as:
\begin{equation}
    F^{l+1} =\phi^l(Sign(W^l) \otimes Sign(F^l))\\
\end{equation}
\begin{equation}
   Sign(x)=\left\{
    \begin{aligned}
+1 & , & x>0, \\
-1 & , & x\le 0.
\end{aligned}
\right.
\end{equation}
where $\otimes$ denotes convolutional operation, the $\phi^l(\cdot)$ denotes the nonlinear operation in the \textsl{l}-layer such as ReLU, PReLU, and BN layer. To reduce the memory saving and computational cost, the BNNs constrain the $W^l$ and $F^l$ to -1 or +1 by Sign function.

\begin{figure}[ht]
    \centering
    \includegraphics[width=\linewidth]{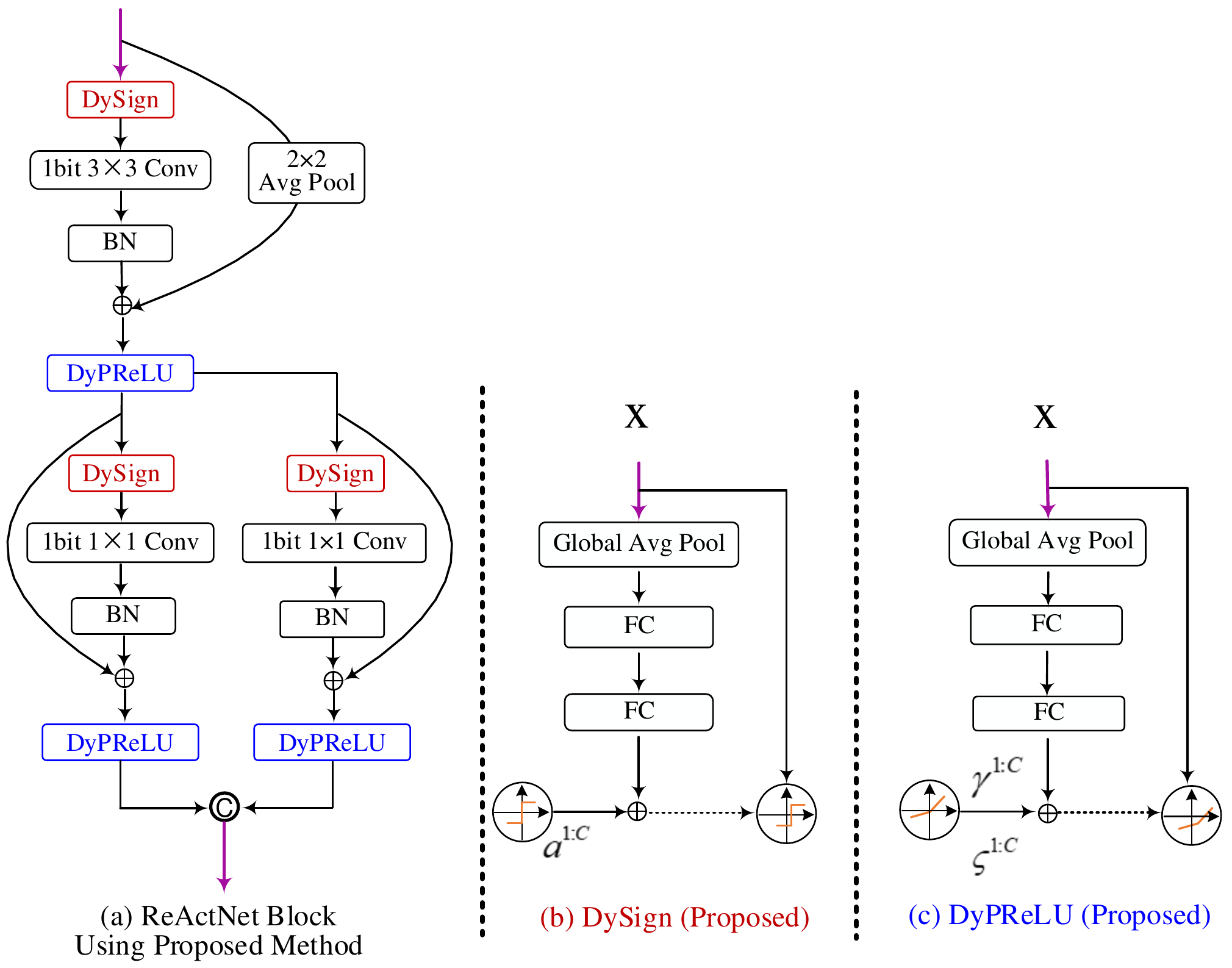}
    \caption{The basic modules of proposed DyBNN (built on the ReActNet).}
    \label{fig:Fig2}
\end{figure}

However, BNNs are sensitive to distribution shift since the activations are constrained to +1 and -1. A slight distribution shift in the input feature map can significantly affect the output of Sign function. To address this problem, the ReActNet \cite{reactnet2020} utilized learnable channel-wise thresholds to shift the feature maps, which is defined as RSign:

\begin{equation}
   RSign(x_i)=\left\{
    \begin{aligned}
+1 & , & x_{i}>a_{i}, \\
-1 & , & x_{i}\le a_{i}.
\end{aligned}
\right.
\end{equation}
where $x_{i}$ denotes the element of \textsl{i}-th channel in input feature map, $a_{i}$ denotes the threshold of  \textsl{i}-th channel, which illustrates that the threshold can vary for different channels. The PReLU function is also processed by this operation to reshape the feature map, which can be expressed as:

\begin{equation}
   RPReLU(x_i)=\left\{
    \begin{aligned}
x_{i}-\gamma_{i}+\zeta_{i} & , & x_{i}>\gamma_{i}, \\
\beta_{i}(x_{i}-\gamma_{i})+\zeta_{i} & , & x_{i}\le \gamma_{i}.
\end{aligned}
\right.
\end{equation}
where $\gamma_{i}$ and $\zeta_{i}$ denote the learnable shift parameters for reshaping the distribution. By introducing the activation distribution shift and reshape, ReActNet achieves a significant increase in image classification.

\subsection{Dynamic binary neural network}

Although ReActNet achieves remarkable performance, the feature information loss still negatively affects the performance of the BNNs since the RSign adopts a \textbf{static} channel-wise thresholds on the input feature maps. Each sample contains a specific feature expression, which means that fixed thresholds cannot be adapted to a broad range of samples (See Fig \ref{fig:Fig1}). The threshold parameters should be adaptively generated to shift the feature maps based on the characteristics of feature. 

Based on the aforementioned statement, we propose the Dynamic Learning Sign Function ($DySign$) to shift the feature maps. For a given input feature map $X$ with $C$ channels, the $DySign$ is defined as a function with learnable parameters $f(x)$, which adapts to the input $X$ (See Fig \ref{fig:Fig2}). The hyper function $f(x)$ computes the threshold for each \textsl{i}th-channel feature $X^i$, which can be expressed as:

\begin{equation}
    \alpha=f(X)=f_2(f_1(\frac{1}{HW}\sum_{H,W}X)) \\
\end{equation}
where $H$ and $W$ denote the height and width of input feature map $X$, $f_1\in \mathbb{R}^{C\times \frac{C}{16}}$ and $f_2\in \mathbb{R}^{\frac{C}{16}}\times C$ denote two fully-connected (FC) layers. The $DySign$ binarized the feature maps based on the parameters $\alpha^{1:C}$ from $f(x)$ as following: 
\begin{equation}
   DySign(x_i)=\left\{
    \begin{aligned}
+1 & , & x_{i}>\alpha_{i}, \\
-1 & , & x_{i}\le \alpha_{i}.
\end{aligned}
\right. \\ \alpha_{i}\in \alpha^{1:C} 
\end{equation}
where $\alpha^{i}$ denotes the threshold of \textsl{i}-th channel, which is the \textsl{i}-th element of output vector from $f(x)$. The $DySign$ adopts a SEBlock \cite{senet2018} to learn a set of channel-wise thresholds from the input feature maps for Sign function. The block of $DySign$ can be observed in Fig \ref{fig:Fig2}. The process is ``$input\to GAP\to FC  layer\to FC layer $''. To avoid over-fit and reduce extra computational cost, the reduction ratio between two linear layers is set as 16. This hyper function encodes the global information of input $F$ to determine appropriate thresholds of each channel. $DySign$ reduces the feature information loss in BNNs and enables significantly more representation power than using standard Sign function.  

We also handle the RPReLU function \cite{reactnet2020} in the same way. The shift parameters $\gamma^{1:C}$ and $\zeta^{1:C}$ are learnable based on input feature maps ($DyPReLU$). The hyper function $f(\cdot)$ generates the shift parameters for each channel in input feature maps. This process can dynamically shift and reshape feature the distribution, which is an effective and simple way to increase model capacity. 

Essentially, the $DySign$ can learn the most suitable thresholds $\alpha^{1:C}$ to binarize the input feature map. The threshold parameters can be dynamically adjusted for different input feature maps, which can effectively limit the feature information loss after binarization. For $DyPReLU$, the $\gamma^{1:C}$ and $\zeta^{1:C}$ can be easily understood as these parameters are dynamically generated to obtain better output distribution. By introducing these functions, the aforementioned problem risen by static parameters can be eliminated. The BNNs can retain more object information and learn more meaningful features. In the experiment section, we will show that dynamic learning distribution is an effective and straightforward way to boost the performance of BNNs. 

\subsection{Model architecture}
For model architecture, The ResNet18 \cite{resnet2016} and MobileNetV1 \cite{mobilenet2017} are built as the backbones following ReActNet \cite{reactnet2020}. In the ReActNet (Shown as in Figure \ref{fig:Fig2}), the 3$\times$3 depthwise and 1$\times$1 pointwise convolution are replaced by standard 3$\times$3 and 1$\times$1 convolution. The duplication and concatenation operations are designed for addressing the channel number difference. In our case, we simply replace the RSign and RPReLU functions in ReActNet with our $DySign$ and $DyPReLU$. The ReActNet can be regarded as the baseline in our experiment.
\vspace{-10pt}
\subsection{Computational complexity analysis}
Following the calculation method in \cite{real2019,reactnet2020}, we calculate total operations (OPs), which consists of binary operations (BOPs) and floating point operations (FLOPs). The OPs can be obtained as:
\begin{equation}
    OPs=\frac{BOPs}{64}+FLOPs \\
\end{equation}
For our DyBNN, we do not introduce extra binary convolutional operations. Thus, the BOPs is same as ReActNet. The increased computational consumption mainly comes from floating-point operations in $DySign$ and $DyPReLU$, including one global average pooling layer and two fully-connected layers. For reducing introduced computational cost, we set the reduction ratio between two fully-connected layers as 16, which can limit the introduced model parameters and float operations. We denote the size of input feature map as $C\times H \times W$. The FLOPs for each RSign and RPReLU increase $C+\frac{C^2}{8}$ and $2\times(C+\frac{C^2}{8})$. \textsl{The extra computational cost is small compared with the total cost.} 

\section{Experiment}
To evaluate the performance of dynamic distribution learning on BNNs, we conduct experiments on ImageNet dataset \cite{imagenet2015}. In this section, we first introduce the training dataset and details. We then report the accuracy and OPs of DyBNN and compare them with state of the art methods. We analyze the impact of $DySign$ and $DyPReLU$ in the ablation study shown in Section \ref{sec:impact}. 

\subsection{Datasets and implementation details}
\label{Sec:training details}
The ILSVRC12 ImageNet classification dataset \cite{imagenet2015} is utilized to evaluate proposed method. The ILSVRC12 ImageNet classification dataset contains 1.2 million training images and 50,000 validation images across 1000 classes, which is more challenging than small datasets.  

 The training strategy utilizes the two-step training strategy described in \cite{real2019}. In the first step, the network is trained from scratch with binary activations and real-valued weights. In the second step, the network inherits the weight from the first step and trained with binary activations and weights. For both steps, we follow the training scheme in \cite{reactnet2020}. We use Adam optimizer and a linear learning rate decay to optimize model. The initial learning rate is 5e-4, and batchsize is set to 256. We also train a quick version of DyBNN with the one-step training strategy. 
 \vspace{-10pt}
\subsection{Experiment on ImageNet}
We compare the DyBNN with other state-of-the-art binarization methods in Table \ref{Tab:1}. Compared with ReActNet and ReActNet-ResNet18, the DyBNN and DyBNN\_ResNet18 achieve 1.8\% and 1.5\% increase, respectively. Due to the introduced global average pooling layer and fully-connected layer, the FLOPs increases slightly. For DyBNN, the OPs is $0.02\times10^8$ higher than ReAcNet (See Tabel \ref{Tab:2}), which is acceptable. With the limited computational cost increased, the DyBNN can outperform previous methods by a large margin, which illustrates the effectiveness of dynamic learnable shift parameters in BNNs. 
\vspace{-15pt}
\begin{table}[htbp]
\centering
\caption{Compare of the top-1 accuracy with state-of-art methods. The W/A denote the number of bits in weight and activation quantization. The \textcolor{blue}{blue} font denotes the comparing result with DyBNN and ReActNet based on ResNet18. The \textcolor{red}{red} font denotes the comparing result based on MobileNetV1.}
\label{Tab:1}
\resizebox{\linewidth}{40mm}{
\begin{tabular}{cccc}
\hline
Binary Method      & W/A & \begin{tabular}[c]{@{}c@{}}Acc Top-1 \\ (\%)\end{tabular} & \begin{tabular}[c]{@{}c@{}}Acc Top-5 \\ (\%)\end{tabular} \\ \hline
BNN\cite{bnn2016}                & 1/1 & 42.2                                                      & 67.1                                                      \\
ABC-Net\cite{abcnet2017}            & 1/1 & 42.7                                                      & 67.6                                                      \\
XNOR-Net\cite{xnornet2016}           & 1/1 & 51.2                                                      & 69.3                                                      \\
DoReFa\cite{dorefa2016}             & 1/2 & 53.4                                                      & -                                                         \\
Bi-RealNet-18\cite{birealnet2018}      & 1/1 & 56.4                                                      & 79.5                                                      \\
XNOR++\cite{xnornet++2019}             & 1/1 & 57.1                                                      & 79.9                                                      \\
IR-Net\cite{irnet2020}             & 1/1 & 58.1                                                      & 80.0                                                      \\
BONN\cite{bayesian2020}               & 1/1 & 59.3                                                      & 81.6                                                      \\
NoisySupervision\cite{noise2020}   & 1/1 & 59.4                                                      & 81.7                                                      \\
RBNN\cite{rotation2020}               & 1/1 & 59.8                                                      & 81.9                                                      \\
Real-to-Binary Net\cite{real2019} & 1/1 & 65.4                                                      & 86.2                                                      \\
\textcolor{blue}{ReActNet-ResNet18}\cite{reactnet2020}  & 1/1 & 65.9                                                      & 86.5                                                      \\
\textcolor{red}{ReActNet}\cite{reactnet2020}           & 1/1 & 69.4                                                      & 88.6                         \\                      AdamBNN\cite{adam2021} & 1/1 & 70.5 & 89.1  \\ \hline
\textcolor{blue}{DyBNN-ResNet18(ours)}     & 1/1 & \textcolor{blue}{67.4$(\uparrow1.5)$}                                                      & 87.4                                                      \\
\textcolor{red}{DyBNN(ours)}              & 1/1 & \textcolor{red}{71.2$(\uparrow1.8)$}                                                      & 89.8                                                      \\
                             \hline
\end{tabular}}
\end{table}
\vspace{-20pt}
\subsection{One-step Training}
\label{sec:onestep}
\begin{figure}[!ht]
    \centering
    \includegraphics[width=0.8\linewidth]{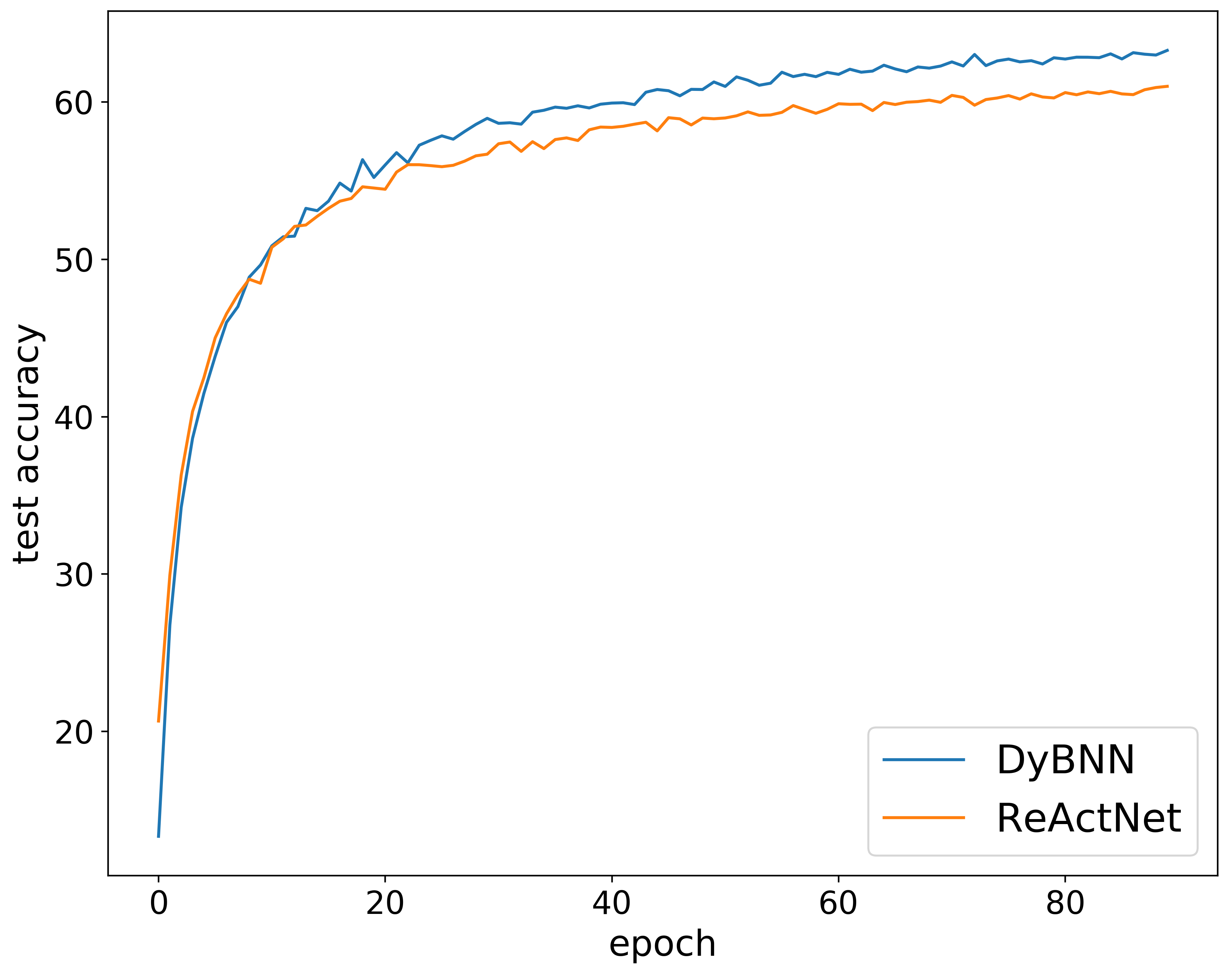}
    \caption{The accuracy of DyBNN and ReActNet on ImageNet.}
    \label{fig:Fig3}
\end{figure}

The DyBNN utilizes the two-step training strategy with 512 epochs in total, which is time-consuming. To simplify the training process, we also evaluate the quick version of DyBNN in one-step training. The result can be observed in Tabel \ref{Tab:2}. We evaluate DyBNN under following strategies: 1) distillation with distribution loss in \cite{reactnet2020}; 2) no distillation with CrossEntropy loss. Following other training details described in \ref{Sec:training details}, DyBNN achieved 2.3\% and 2.0\% higher than ReActNet under the two training strategies mentioned above, which illustrated that DyBNN is an effective and straightforward way to boost performance of BNNs. We also observe that DyBNN achieves higher accuracy and faster convergence (See Fig \ref{fig:Fig3}). In this paper, The calculation of FLOPs contains BN, PReLU layers, so our reported OPs is higher than $0.87\times10^8$ in \cite{reactnet2020}.

\begin{table}[htbp]
\centering
\caption{The test result of a quick version for DyBNN. The \Checkmark denotes the model is trained with distributional loss and the real-valued ResNet34 is set as teacher model. No \Checkmark denotes the model is directly trained with Cross Entropy loss.}
\label{Tab:2}
\begin{tabular}{cccc}
\hline
Network  & distillation & OPs                       & Acc Top-1(\%) \\ \hline
ReActNet &              & \multirow{2}{*}{$0.97\times 10^8$} &  61.0             \\
ReActNet &    \Checkmark          &                           & 64.3              \\ \hline
DyBNN    &              & \multirow{2}{*}{$0.99\times 10^8$} & \textbf{63.3}          \\
DyBNN    &    \Checkmark         &                           & \textbf{66.3}            \\ \hline
\end{tabular}
\end{table}

\subsection{Impacts of DySign and DyPReLU}
\label{sec:impact}
In this section, we analyze the individual effect of $DySign$. We conduct the experiment in the two-step training strategy following Section \ref{Sec:training details}. The experimental result is shown in Table \ref{Tab:3}. DyBNN achieves 70.1\% top-1 accuracy on the ImageNet without $DyPReLU$. Comparing with ReActNet, the accuracy increases by 0.7\%, illustrating the effectiveness of our proposed $DySign$. When adding $DyPReLU$, DyBNN can achieve 71.2\% top-1 accuracy. Shown in Fig \ref{fig:Fig1}, $DySign$ promotes model performance by reducing feature information loss. Furthermore, $DyPReLU$ increases model expression ability by reshaping activation distribution of feature maps, which further improves model accuracy.
\vspace{-18pt}
\begin{table}[ht]
\centering
\caption{The impact of \textsl{DySign} and \textsl{DyPReLU}.}
\label{Tab:3}
\begin{tabular}{cccc}
\hline
Network & DySign & DyPReLU & Acc Top-1(\%) \\ \hline
ReActNet  &                 &         &    69.4         \\
DyBNN   &   \Checkmark     &         &    70.1           \\
DyBNN   & \Checkmark     &    \Checkmark     &   71.2            \\\hline
\end{tabular}
\end{table}
\vspace{-20pt}
\section{Conclusion}
\vspace{-10pt}
In this paper, we have introduced the dynamic learning method into binary neural networks and propose DyBNN. DyBNN dynamically generates adaptive parameters to shift and reshape distribution activations. The experimental results show that DyBNN can reduce the feature information loss and significantly improve the performance of BNNs with a limited computational cost increase. 

\bibliographystyle{IEEE}
\end{document}